\documentclass[journal]{IEEEtran}

\ifCLASSINFOpdf
  \usepackage{graphicx}
  \graphicspath{{figures/}{../figures/}{./}}
\else
  \usepackage[dvips]{graphicx}
  \graphicspath{{figures/}{../figures/}{./}}
\fi

\usepackage[utf8]{inputenc}
\usepackage[T1]{fontenc}
\usepackage{url}
\usepackage{cite}
\usepackage{amsmath,amssymb,amsfonts}
\usepackage{textcomp}
\usepackage{xcolor}
\usepackage{booktabs}
\usepackage{microtype}
\usepackage{bm}
\usepackage{multirow}
\usepackage[hidelinks]{hyperref}

\providecommand{\R}{\mathbb{R}}

\begin{document}

\title{Groupwise Selective State-Space Filtering for Accurate and Streaming Action Boundary Detection}

\author{Mustafa Bora \protect\c{C}elik%
\thanks{This work has been submitted to the IEEE for possible publication. Copyright may be transferred without notice, after which this version may no longer be accessible.}%
\thanks{The author is with the Department of Electrical and Electronics Engineering, Ankara Medipol University, Ankara, Turkey (e-mail: mustafa.celik1@std.ankaramedipol.edu.tr).}%
\thanks{Code and reproducible models are publicly available at \protect\href{https://github.com/BOA-clk/Boundary_Detection}{\nolinkurl{https://github.com/BOA-clk/Boundary_Detection}}.}}

\markboth{Preprint. Under review.}%
{\protect\c{C}elik: Groupwise Selective State-Space Filtering for Action Boundary Detection}

\maketitle

\begin{abstract}
Action boundary detection partitions untrimmed video into intervals without assigning action classes. We present a boundary-detection adapter operating on pre-extracted video features, learning temporal representations via groupwise selective scans. Learned group fusion and temporal modeling convert these into transition scores, which are decoded into boundary timestamps. Trained with boundary-time supervision, the class-agnostic model is evaluated on Breakfast, GTEA, and 50Salads using temporal tolerances and bipartite matching, achieving boundary $F_1$ scores of 0.457, 0.622, and 0.611. A stateful variant enables feature-streaming inference with zero neural look-ahead, one-sample peak confirmation, and bounded memory. Downstream systems can subsequently assign semantics to these detected temporal units.
\end{abstract}


\IEEEpeerreviewmaketitle

\section{Introduction}
\IEEEPARstart{L}{ocating} action transitions provides temporal units for processing long, untrimmed videos. Temporal action segmentation usually combines this task with assigning an action class to every frame \cite{farha2019ms,ishikawa2021action,yi2021asformer,baformer2024,diffact2023}. A boundary-detection adapter can instead expose candidate intervals without assigning semantic labels to them, allowing a downstream retrieval or video-language model (VLM) to interpret their content. This is a potential application rather than an evaluated downstream capability: we assess boundary localization on Breakfast, GTEA, and 50Salads, whose action annotations provide reference transition times. The detector uses these times during training; class-agnostic output does not imply unsupervised learning.

Evaluation should directly reflect timestamp output rather than frame-level action classes or segmental edit scores. We evaluate predicted and reference boundaries within fixed temporal tolerances via bipartite one-to-one matching: a matched pair within tolerance contributes one true positive, whereas unmatched references and redundant predictions contribute false negatives and false positives, respectively. Boundary precision, recall, and $F_1$ thus quantify localization independently of action classes under narrow, primary, and wide tolerances (Section~\ref{sec:protocol}). This protocol complements conventional segmentation metrics; generic event boundary detection studies broader taxonomy-free changes \cite{shou2021gebd}, whereas our references are strictly action transitions.

The adapter is built around independent selective state-space scans. Fixed feature subspaces are aligned into isolated temporal chains, allowing every chain to carry its own recurrent state through the video. Their state-mediated outputs are fused into a compact transition representation, enriched with local temporal context, and mapped to boundary scores before structured timestamp decoding. Because the underlying scan is recurrent, the same formulation also supports a stateful feature-streaming realization: persistent states are updated once for each arriving feature, without recomputing the preceding sequence or growing a history cache.
\begin{figure}[t]
\centering
\IfFileExists{figures/model_overview.pdf}{%
  \includegraphics[width=0.98\columnwidth]{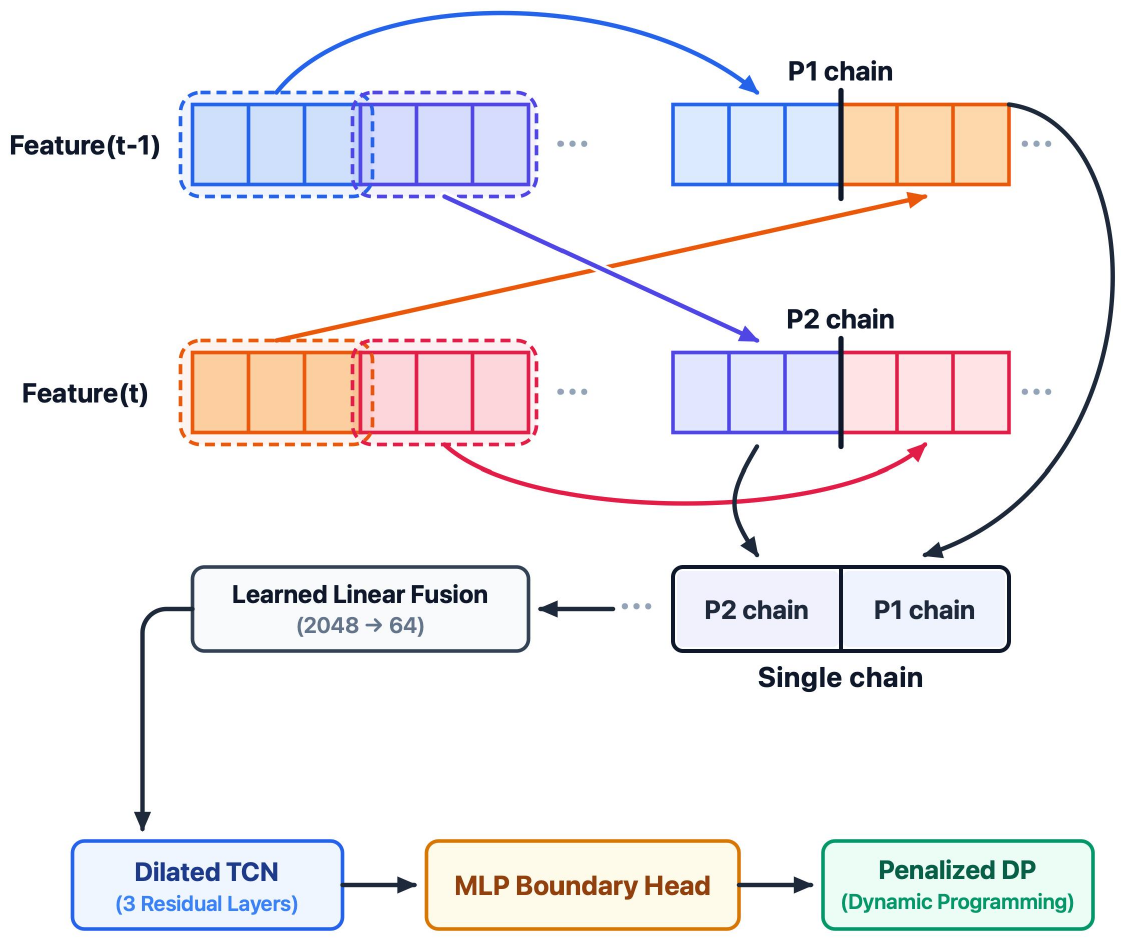}%
}{%
  \includegraphics[width=0.98\columnwidth]{../figures/model_overview.pdf}%
}
\caption{The proposed architecture aligns feature subspaces into isolated temporal group chains ($P_1$ and $P_2$) processed by independent Mamba blocks without cross-group state sharing. At each time step, their recurrent outputs are concatenated along the feature dimension, projected via learned linear fusion ($2048 \to 64$), and processed by a dilated TCN, MLP boundary head, and penalized dynamic programming (DP) decoder.}
\label{fig:overview}
\end{figure}
Our contributions are:
\begin{itemize}
 \item A boundary-detection adapter combining isolated groupwise selective scans, learnable group fusion, dilated temporal context, and structured decoding. Its full pipeline achieves the best mean primary boundary $F_1$ among the completed baseline reproductions on Breakfast, GTEA, and 50Salads.
 \item A fully stateful feature-streaming realization that provides constant per-step computation and bounded $O(1)$ recurrent memory with respect to video duration, operating with zero neural look-ahead without recomputing the observed prefix.
\end{itemize}

\section{Method}
\subsection{From Interleaved Scanning to Temporal Group Chains}
Let $X=(x_0,\ldots,x_{T-1})$, with $x_t\in\R^{2048}$, denote the 15-Hz sequence of pre-extracted I3D features \cite{carreira2017quo}. Reference boundaries $\mathcal G=\{g_j\}_{j=1}^{K}$ are timestamps at which consecutive action annotations differ. No temporal averaging is applied.

Our design adopts the scan principle of MamBOA \cite{celik2026mamboa}, which absorbs temporal relations into state dynamics by scanning corresponding feature subspaces consecutively. For multi-transition localization in long I3D sequences, we adapt this principle by replacing paired patch interleaving with persistent, mutually isolated channel-group chains, ensuring corresponding subspaces remain adjacent along the temporal scan without cross-group leakage.

Each feature is split into $G=32$ contiguous groups,
\begin{equation}
 x_t=[u_t^{(1)};\ldots;u_t^{(G)}],\qquad u_t^{(g)}\in\R^{64}.
 \label{eq:groups}
\end{equation}
The usual time-major layout places all groups of $x_t$ next to each other. We re-index it into $G$ aligned temporal chains,
\begin{equation}
\begin{aligned}
 \mathcal S^{(1)} &: u_0^{(1)}\rightarrow u_1^{(1)}\rightarrow\cdots\rightarrow u_{T-1}^{(1)},\\
 \mathcal S^{(2)} &: u_0^{(2)}\rightarrow u_1^{(2)}\rightarrow\cdots\rightarrow u_{T-1}^{(2)},\\[-2pt]
 &\hspace{14mm}\vdots\\[-2pt]
 \mathcal S^{(G)} &: u_0^{(G)}\rightarrow u_1^{(G)}\rightarrow\cdots\rightarrow u_{T-1}^{(G)}.
\end{aligned}
\label{eq:chains}
\end{equation}
Thus, as illustrated in Fig.~\ref{fig:overview}, each recurrent state is updated only by the same feature subspace observed at successive times; information cannot leak between groups through a shared recurrent state. Because the I3D descriptor used here has no explicit spatial grid, $u_t^{(g)}$ is a channel group rather than a literal image patch. The connection to \cite{celik2026mamboa} is structural: both scan orders align corresponding subspaces across observations.

Each chain is processed by an independent forward Mamba block \cite{gu2024mamba}. Layer normalization is applied within the current 64-dimensional group and never across time. Let $v_t^{(g)}$ denote the group feature after the block's input projection and causal local convolution, and let $a_t^{(g)}$ denote its multiplicative gate. The selective state-space core contains a state-mediated term and an additive input bypass,
\begin{equation}
 \widetilde o_t^{(g)}=C_t^{(g)}h_t^{(g)}+D^{(g)}v_t^{(g)}.
 \label{eq:readout}
\end{equation}
The complete Mamba output also retains its multiplicative gate and output projection. We freeze $D^{(g)}=0$ and compute
\begin{equation}
 o_t^{(g)}=C_t^{(g)}h_t^{(g)},\qquad
 m_t^{(g)}=W_o^{(g)}\!\left(o_t^{(g)}\odot\operatorname{SiLU}(a_t^{(g)})\right).
 \label{eq:no_skip}
\end{equation}
This removes the shortest additive appearance-only path, while preserving Mamba's input-dependent gate; it neither exposes the latent state directly nor forces $m_t^{(g)}$ to equal a hand-crafted finite difference. We use state dimension 16, local convolution width 4, and expansion factor 1; the $G$ Mamba blocks do not share recurrent state.

The arrows in Eq.~\eqref{eq:chains} denote an actual recurrent state chain rather than independent per-frame transformations. For group $g$, starting from $h_{-1}^{(g)}=0$, the state is written sequentially as
\begin{equation}
 h_t^{(g)}=
 \bar A_t^{(g)}h_{t-1}^{(g)}+
 \bar B_t^{(g)}v_t^{(g)}
 =\sum_{j=0}^{t}
 \left(\prod_{k=j+1}^{t}\bar A_k^{(g)}\right)
 \bar B_j^{(g)}v_j^{(g)}.
 \label{eq:state_chain}
\end{equation}
The product in Eq.~\eqref{eq:state_chain} is time ordered with $\bar A_t^{(g)}$ leftmost; for $j=t$ it is the identity.
Consequently, $h_t^{(g)}$ is not computed and discarded at time $t$: after producing $m_t^{(g)}$, it becomes the memory from which $h_{t+1}^{(g)}$ is formed. The input-dependent transition and write terms determine how much of the previous state is retained and how strongly the current group feature is incorporated. When a feature similar to the established group pattern reappears, the model can learn a small effective write and largely preserve its memory; when the incoming pattern changes and remains changed, successive updates rewrite the state. Reading out every state $h_0^{(g)},\ldots,h_{T-1}^{(g)}$ therefore yields a trajectory that follows the frame sequence rather than a collection of unrelated frame descriptors. A parallel scan kernel may evaluate this recurrence efficiently during training, but it does not change these sequential state semantics.

\subsection{Group Fusion and Boundary Formation}
At each time step, the $G$ isolated scan outputs are restored to their original group order and concatenated. A learned projection reduces the resulting 2048-dimensional state-mediated descriptor to a compact boundary representation,
\begin{equation}
 r_t=[m_t^{(1)};\ldots;m_t^{(G)}],\qquad
 z_t=W_f r_t+b_f,\quad z_t\in\R^{64}.
 \label{eq:fusion}
\end{equation}
Unlike mean or norm pooling, this projection can preserve which groups respond and learn how their responses should be combined from boundary supervision.

The sequence $z_{0:T-1}$ is then processed by three residual TCN layers. Each layer applies LayerNorm, a width-3 dilated convolution, SiLU, and a residual addition. We use dataset-specific dilation triplets of $(1,11,23)$, $(1,7,13)$, and $(1,17,35)$ for Breakfast, GTEA, and 50Salads, respectively. Symmetric padding allows every boundary score to use local evidence on both sides of a candidate transition. A learnable $64\!\rightarrow\!64\!\rightarrow\!1$ MLP maps the contextual representation to the boundary logit $s_t$.

Reference boundaries supervise a soft target
\begin{equation}
 y_t=\max_{g_j\in\mathcal G}\exp\!\left[-\frac{(t/15-g_j)^2}{2\sigma^2}\right],
 \label{eq:target}
\end{equation}
where $\sigma$ is 0.8, 0.3, and 1.0~s for Breakfast, GTEA, and 50Salads. We minimize positive-weighted binary cross entropy,
\begin{equation}
 \mathcal L=\frac{\sum_t(1+\alpha y_t)\operatorname{BCEWithLogits}(s_t,y_t)}
 {\sum_t(1+\alpha y_t)},\qquad \alpha=4.
 \label{eq:loss}
\end{equation}

The soft target accommodates slight timing deviations around annotations. At inference, logits are standardized per video (std floor 0.5) to obtain scores $q_t$. Penalized dynamic programming (DP) selects timestamps by maximizing cumulative scores with a separation constraint $m$ and penalty $\lambda$:
\begin{equation}
 V(t)=\max\{V(t-1),\;V(t-m)+q_t-\lambda\}.
 \label{eq:dp}
\end{equation}
Backtracking extracts boundary indices, followed by three-point parabolic sub-step refinement and validation-calibrated temporal bias correction. Streaming inference replaces DP with causal peak confirmation.
\begin{figure*}[t]
\centering
\IfFileExists{figures/action_segmentation_matplotlib.pdf}{%
  \includegraphics[width=0.98\textwidth]{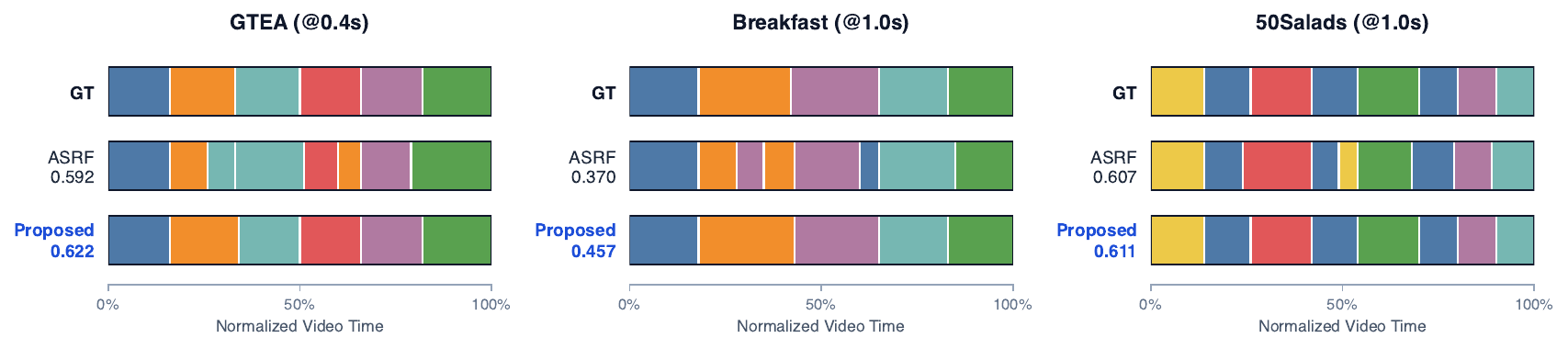}%
}{%
  \includegraphics[width=0.98\textwidth]{../figures/action_segmentation_matplotlib.pdf}%
}
\caption{Qualitative comparison of action boundary predictions on representative sequences from GTEA, Breakfast, and 50Salads against Ground Truth (GT) and ASRF. Scores below method names indicate primary $F_1$ boundary detection performance.}
\label{fig:boundary_timelines}
\end{figure*}

\subsection{Stateful Feature Streaming}
The TCN strengthens the boundary representation by comparing evidence on both sides of a candidate, but its symmetric temporal convolutions require future features. For strictly sequential operation, we train a separate TCN-free configuration with the boundary head applied directly to $z_t$. The groupwise Mamba formulation is unchanged. Each block retains its local-convolution buffer and state-space state, and an arriving $x_t$ updates every group exactly once; neither the complete prefix nor a growing feature cache is recomputed.

Video-global score normalization is replaced by Welford running statistics. Once step $t$ arrives, the decoder can decide whether the previous score $q_{t-1}$ is a local maximum above its validation-selected threshold. The Mamba representation therefore has zero neural look-ahead, while local-peak confirmation introduces one sample of decision look-ahead.
The streaming interface begins when $x_t$ becomes available; causality of the upstream I3D extractor is outside the model defined here.

\section{Experiments}
\subsection{Protocol}
\label{sec:protocol}
We evaluate on the official splits of Breakfast (4 folds), GTEA (4 folds), and 50Salads (5 folds) \cite{kuehne2014breakfast,fathi2011understanding,stein2013salads} using identical 2048-dimensional features. Following standard preprocessing, features are subsampled with factor $s=2$ on 50Salads and $s=1$ on Breakfast/GTEA; reference and predicted transition timestamps are placed at $(i-0.5)/f_{\mathrm{src}}$ and $(j-0.5)s/f_{\mathrm{src}}$, respectively. Alignment checks ensure reference timestamps match within $10^{-9}$~s.

Each fold is trained with seeds 0 and 1 (20\% validation split). Baselines use Adam ($\text{lr}=5\!\times\!10^{-4}$, $\le$50 epochs, patience 10, batch size 1). Our adapter uses AdamW at the same initial rate for up to 60 epochs with weight decay 0.01, 10\% linear warm-up, cosine decay, and patience 10. Validation $F_1$ at the primary tolerance governs checkpointing and baseline post-processing. A single validation-calibrated temporal offset (median signed residual) is subtracted from test predictions without modifying training.

A prediction and reference are matched at most once by maximum-cardinality, minimum-error bipartite assignment. Tolerances are $(0.5,1,2)$~s for Breakfast and 50Salads and $(0.2,0.4,0.8)$~s for GTEA; the middle tolerance is primary. We report video-macro $F_1$. The stated variability is the standard deviation across fold/seed runs and is not treated as an independent-sample confidence interval.
\begin{table}[h]
\centering
\caption{Primary action-boundary $F_1$ under the common one-to-one protocol. Bold and underlined entries are best and second best. N/C: reproduction not completed within the available compute budget.}
\label{tab:main}
\footnotesize
\setlength{\tabcolsep}{3.2pt}
\renewcommand{\arraystretch}{1.12}
\begin{tabular*}{\columnwidth}{@{\extracolsep{\fill}}l r ccc@{}}
\toprule
Method & Params & Breakfast & GTEA & 50Salads \\
& & $@1$~s & $@0.4$~s & $@1$~s \\
\midrule
MS-TCN \cite{farha2019ms}      & 0.80M & 0.349 & 0.557 & 0.370 \\
ASFormer \cite{yi2021asformer} & 1.13M & N/C   & 0.575 & 0.455 \\
BaFormer \cite{baformer2024}   & 1.63M & 0.313 & \underline{0.606} & 0.524 \\
DiffAct \cite{diffact2023}     & 1.21M & 0.182 & 0.285 & 0.312 \\
ASRF \cite{ishikawa2021action} & 1.30M & \underline{0.370} & 0.592 & \underline{0.607} \\
\textbf{Ours}                  & \textbf{0.70M} & \textbf{0.457} & \textbf{0.622} & \textbf{0.611} \\
\bottomrule
\end{tabular*}
\end{table}

\subsection{Boundary-Detection Results}
Table~\ref{tab:main} reports the primary boundary score for each dataset under our one-to-one protocol, rather than the original semantic-segmentation scores of the cited papers. The ASFormer Breakfast reproduction is marked N/C because it did not finish within the common compute budget; no value is imputed. At the primary tolerance, the full pipeline improves over the strongest evaluated literature baseline by 0.087, 0.016, and 0.004 absolute $F_1$ on Breakfast, GTEA, and 50Salads, respectively. Its corresponding run standard deviations are 0.016, 0.030, and 0.046. The 50Salads margin is small relative to this variability and is interpreted as the best mean rather than an established significant advantage. Qualitative boundary predictions on representative sequences from all three datasets are visualized in Fig.~\ref{fig:boundary_timelines}. In the illustrated sequences, our adapter aligns closely with ground-truth transitions and suppresses ASRF's redundant predictions within uniform action intervals.

\begin{table}[t]
\centering
\caption{Stateful streaming results on 50Salads: five folds, two seeds, zero neural look-ahead, and one-sample peak confirmation. Width denotes channels per recurrent group.}
\label{tab:scan_granularity}
\footnotesize
\setlength{\tabcolsep}{3.0pt}
\renewcommand{\arraystretch}{1.05}
\begin{tabular*}{\columnwidth}{@{\extracolsep{\fill}}rrc@{}}
\toprule
Width & Params & $F_1@1$ \\
\midrule
2048 & 13.359M & $0.4687{\pm}0.0735$ \\
64   & 0.658M  & $\mathbf{0.4818{\pm}0.0653}$ \\
32   & \textbf{0.453M} & $0.4736{\pm}0.0460$ \\
\bottomrule
\end{tabular*}
\end{table}

\subsection{Streaming Results and Group Width}
Table~\ref{tab:scan_granularity} reports streaming boundary-detection results on 50Salads for three group widths. Each configuration is trained without the TCN and processes one arriving feature at a time using persistent recurrent states, running score normalization, and causal peak confirmation. Neural look-ahead is zero; the decoder confirms a boundary after one subsequent feature arrives.

The configuration with group width 2048 scans the complete descriptor using one recurrent state-space block. A group width of 64 produces 32 independent temporal chains, whereas a group width of 32 produces 64 chains. The narrower projections reduce parameter counts by factors of $20.3$ and $29.5$ relative to processing the complete descriptor in one block. We use group width 64 in the proposed adapter: it achieves the highest mean streaming score in this comparison with 0.658M parameters. The score differences remain small relative to run variability, so this experiment supports parameter efficiency rather than an established significant accuracy gain. All three configurations retain 160~KiB of FP32 convolution and state-space state. For the proposed width-64 configuration, the measured 50Salads implementation processes features at 50.5 steps/s with a p95 processing time of 20.45~ms per feature, supporting the 15-Hz input rate. This processing time excludes waiting for the next feature required by peak confirmation and excludes upstream I3D extraction.
\subsection{Computational Complexity and Streaming Invariants}
\begin{figure}[t]
\centering
\IfFileExists{figures/flop_trend.pdf}{%
  \includegraphics[width=\columnwidth]{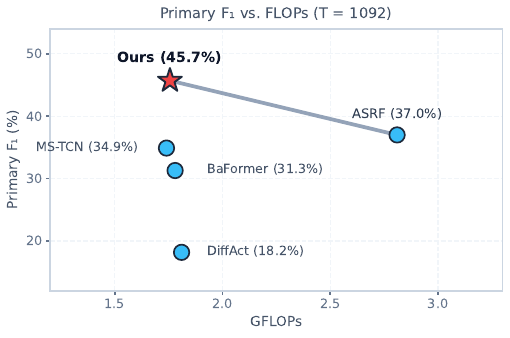}%
}{%
  \includegraphics[width=\columnwidth]{../figures/flop_trend.pdf}%
}
\caption{Primary boundary $F_1$ versus analytically estimated complexity at reference sequence length $T=1092$ on Breakfast. The proposed model ($\bigstar$) attains $45.7\%$ $F_1$ at approximately 1.756~GFLOPs, compared with MS-TCN (1.740~GFLOPs, $34.9\%$) and ASRF (2.810~GFLOPs, $37.0\%$).}
\label{fig:flop_trend}
\end{figure}

Fig.~\ref{fig:flop_trend} compares primary $F_1$ on Breakfast against analytically estimated computational complexity at reference sequence length $T=1092$. Baselines MS-TCN \cite{farha2019ms} (1.740~GFLOPs, 34.9\% $F_1$), BaFormer \cite{baformer2024} (1.780~GFLOPs, 31.3\%), and DiffAct \cite{diffact2023} (1.810~GFLOPs, 18.2\%) cluster at low estimated FLOPs. Our full pipeline requires approximately 1.756~GFLOPs (0.70M parameters), giving 10.8 percentage points higher $F_1$ than MS-TCN at a comparable estimated cost. It is also estimated to use $1.60\times$ fewer FLOPs than ASRF \cite{ishikawa2021action} (2.810~GFLOPs, 37.0\%) and more than $2\times$ fewer than ASFormer \cite{yi2021asformer} (3.600~GFLOPs). Because architectures contain different operators, these analytical values are interpreted as consistent implementation-level estimates rather than exact hardware-independent costs.

\textbf{Streaming complexity.} The width-64 configuration requires an estimated 1.534~MFLOPs per feature, or 1.675~GFLOPs for $T=1092$. Our model counts are analytical estimates: a multiplication and an addition count as two FLOPs, and the selective scan uses a $9TEN$ proxy, where $E$ and $N$ denote inner width and state dimension, summed across groups. Normalization, most nonlinearities, biases, residual additions, decoding, data transfer, and I3D extraction are excluded. For fixed model dimensions, updating the 32 recurrent states, running statistics, and peak decoder requires $O(1)$ work per feature with respect to processed video length. Persistent working memory is also $O(1)$, comprising 160~KiB of FP32 convolution/SSM state plus fixed-size normalization and decoder buffers; emitted boundaries can be consumed incrementally. Processing $T$ features therefore takes $O(T)$ total computation without retaining or recomputing the observed prefix.

\subsection{Component Ablation on 50Salads}
Table~\ref{tab:ablation} summarizes component removals and replacements on 50Salads, each evaluated over five folds and two training seeds. The full-pipeline entry is the main experiment from Table~\ref{tab:main}; the component variants come from a separate ablation training batch. Their aggregate scores are descriptive comparisons, not paired estimates of component effects.

\begin{table}[h]
\centering
\caption{Component study on 50Salads (five folds, two seeds). $\dagger$: full-pipeline result from the main experiment; component variants were trained in a separate batch.}
\label{tab:ablation}
\footnotesize
\setlength{\tabcolsep}{2.6pt}
\renewcommand{\arraystretch}{1.05}
\begin{tabular*}{\columnwidth}{@{\extracolsep{\fill}}lc@{}}
\toprule
Configuration & Primary $F_1@1$ \\
\midrule
\textbf{Full pipeline}$^\dagger$ & $\mathbf{0.611{\pm}0.046}$ \\
\midrule
w/o nonlinear head (linear head) & $0.603{\pm}0.032$ \\
w/o input normalization & $0.593{\pm}0.040$ \\
w/o learned fusion (mean pooling) & $0.578{\pm}0.056$ \\
w/o TCN & $0.559{\pm}0.044$ \\
\bottomrule
\end{tabular*}
\end{table}

The reported mean scores are lower when input normalization, learned fusion, or temporal context is removed. Replacing the MLP with a single linear boundary head yields 0.603, while replacing learned fusion with uniform group averaging yields 0.578. The configuration without the TCN has the lowest mean in this table (0.559), consistent with the utility of local two-sided context. This TCN-removal experiment retains the full-pipeline decoding protocol and is distinct from the stateful streaming evaluations in Table~\ref{tab:scan_granularity}.

\section{Discussion and Conclusion}
The proposed adapter localizes action transitions without assigning semantic labels to the resulting intervals. Independent recurrent filtering of feature groups supports both a full pipeline with two-sided context and a stateful streaming configuration with bounded memory. Experiments establish boundary-localization performance on three action benchmarks and measured feature-streaming operation; they do not establish raw-video causality or downstream VLM gains.

\textbf{Future work.} Promising directions include evaluating whether detected intervals improve long-video retrieval or VLM-based reasoning, as well as examining video backbones with longer temporal context to assess their impact on boundary detection.

\newpage
\bibliographystyle{IEEEtran}

\end{document}